\documentclass[conference]{IEEEtran}
\IEEEoverridecommandlockouts    

\usepackage{cite}
\usepackage{multirow}
\usepackage{lipsum}
\usepackage{amsmath,amsfonts}
\usepackage{amssymb}
\usepackage[ruled,vlined]{algorithm2e}
\usepackage{graphicx}
\usepackage{algpseudocode}

\usepackage{multirow}
\usepackage{booktabs}
\graphicspath{{./Figures/}}

\title{\LARGE \bf Beyond Conservative Automated Driving in Multi-Agent Scenarios via Coupled Model Predictive Control and Deep Reinforcement Learning}

\author{
\parbox{\textwidth}{%
\centering
Saeed Rahmani$^{1,*}$, G\"ozde K\"orpe$^{1}$, Zhenlin (Gavin) Xu$^{1}$, Bruno Brito$^{2}$, Simeon Craig Calvert$^{1}$, Bart van Arem$^{1}$%
}%
\thanks{$^{1}$TU Delft, Faculty of Civil Engineering and Geosciences, Department of Transport and Planning, Delft, The Netherlands}%
\thanks{$^{2}$NVIDIA, San Jose, CA, USA}%
\thanks{$^{*}$Corresponding author: {\tt\small s.rahmani@tudelft.nl}}%
\thanks{$^{\dagger}$This work has been submitted to the IEEE for possible publication. Copyright may be transferred without notice, after which this version may no longer be accessible.}%
}

\begin{document}
	
	\maketitle
	\thispagestyle{empty}
	\pagestyle{empty}
	
	\begin{abstract}
    Automated driving at unsignalized intersections is challenging due to complex multi-vehicle interactions and the need to balance safety and efficiency. Model Predictive Control (MPC) offers structured constraint handling through optimization but relies on hand-crafted rules that often produce overly conservative behavior. Deep Reinforcement Learning (RL) learns adaptive behaviors from experience but often struggles with safety assurance and generalization to unseen environments. In this study, we present an integrated MPC-RL framework to improve navigation performance in multi-agent scenarios. Experiments show that MPC-RL outperforms standalone MPC and end-to-end RL across three traffic-density levels. Collectively, MPC-RL reduces the collision rate by 21\% and improves the success rate by 6.5\% compared to pure MPC. We further evaluate zero-shot transfer to a highway merging scenario without retraining. Both MPC-based methods transfer substantially better than end-to-end PPO, which highlights the role of the MPC backbone in cross-scenario robustness. The framework also shows faster loss stabilization than end-to-end RL during training, which indicates a reduced learning burden. These results suggest that the integrated approach can improve the balance between safety performance and efficiency in multi-agent intersection scenarios, while the MPC component provides a strong foundation for generalization across driving environments. The implementation code is available open-source.
	\end{abstract}
	
\section{Introduction}
The rapid advancement of automated driving technology promises to revolutionize transportation by offering increased safety, efficiency, and passenger comfort. A critical component of this technology is the development of robust and adaptable motion planning algorithms, particularly in complex driving scenarios such as unsignalized intersections. These scenarios demand not only sophisticated environmental understanding but also the ability to engage in implicit cooperation or negotiation with human drivers based on implicit rules and social conventions \cite{zhang2021survey, hubmann2017decision}. The social dynamics of driving demand a careful balance between assertive and cooperative behaviors, where insufficient assertiveness causes congestion while excessive aggression compromises safety. 

Various approaches have been proposed to address this tension. Model Predictive Control (MPC) offers a promising approach due to its ability to handle system constraints \cite{de2023model}. However, MPC approaches often rely on pre-defined, hand-crafted rules and cost functions, which leads to overly conservative maneuvers in complex scenarios \cite{brito2022learning, rahmani2023bi}. Deep reinforcement learning (Deep RL) methods have shown promise in learning complex driving behaviors through trial-and-error interactions with the environment, but they often struggle with safety guarantees and generalization to new scenarios \cite{bae2023safe}. 

Recent hybrid approaches have attempted to combine MPC and RL to leverage their complementary strengths. Most of these methods employ switching or supervisory mechanisms in which RL and MPC operate independently; for example, alternating control authority based on safety criteria \cite{wang2020safe, li2020deep}, selecting between parallel RL and MPC planners \cite{bautista2022autonomous}, or using RL solely to decide when MPC should recompute \cite{dang2023event}. Among the few studies that adopt a more integrated architecture \cite{liniger2019safe, zhang2023learning}, the evaluation has been limited to simplified scenarios involving only one or two surrounding vehicles, and the generalizability of the learned policies to structurally different driving environments has not been investigated. Moreover, their training procedure disables the planner's collision avoidance constraints to expose the RL agent to collisions, then re-enables them at deployment~\cite{brito2022learning}. This introduces a mismatch between training and deployment dynamics, which can lead to suboptimal interaction between the two components during operation.

To address these gaps, we propose an integrated MPC-RL framework for safer and more efficient navigation in multi-agent scenarios. In our architecture, RL provides high-level speed guidance, and MPC handles low-level trajectory optimization, with both components coupled at every control timestep rather than operating independently or at different time scales. This stands in contrast to switching-based approaches, where MPC and RL alternate control authority, or supervisory schemes, where one component monitors the other without continuous bidirectional interaction. Furthermore, our framework maintains the MPC's collision avoidance mechanism throughout both training and evaluation. This ensures that the RL agent learns speed references that account for and complement the MPC's reactive safety behavior, thereby avoiding the train-deployment mismatch that arises when the safety layer is absent during learning. We train and evaluate the proposed framework in multi-vehicle, multi-directional intersection scenarios across three traffic-density levels and assess zero-shot transferability to highway merging without retraining. The results show consistent improvements over both standalone MPC and end-to-end RL, with robust zero-shot transfer to merging, especially in less congested scenarios.

Our framework offers several practical advantages. First, the explicit modeling of vehicle dynamics within the MPC component ensures that generated trajectories respect the modeled physical constraints, which supports kinematic feasibility and safety-oriented behavior. Second, by restricting the RL agent to learning high-level speed references rather than the full control policy, the framework significantly reduces the learning burden, which leads to faster loss stabilization during training. Furthermore, the MPC component acts as a consistent control layer that enforces kinematic feasibility and input constraints throughout operation, so that even if the RL agent proposes aggressive speed references, the executed trajectory remains bounded by the MPC optimization and the collision-avoidance logic. The implementation code is omitted for anonymous review and will be released upon publication.

\section{Related Work}
This section reviews the main approaches proposed to address the challenges of automated driving at unsignalized intersections or other comparable interactive scenarios.

\subsection{Model Predictive Control Approaches}
MPC has emerged as a promising method for automated vehicle motion planning due to its ability to handle system constraints and optimize trajectories over a finite prediction horizon \cite{de2023model}. As examples of studies using this approach, \cite{qian2020cooperative} proposed a decentralized MPC approach for cooperative trajectory planning of connected and automated vehicles at intersections. In \cite{wang2021game}, a game-theoretic MPC framework is presented to tackle motion planning during merging maneuvers. Furthermore, an adaptive MPC scheme is introduced in \cite{zhao2017adaptive} to account for changing road conditions, which can be useful in challenging weather conditions. However, a common limitation of traditional MPC approaches is their reliance on pre-defined, hand-crafted rules and cost functions to capture the complexities of driving behavior, often leading to overly conservative maneuvers in complex scenarios where some form of negotiation is needed for resolving potential conflicts.

\subsection{Reinforcement Learning Methods}
RL has gained significant traction in automated driving research in recent years. Several studies have applied deep RL to intersection and multi-agent scenarios. \cite{bae2023safe} proposed an RL-based controller for navigating intersections based on Control Barrier Functions. In \cite{mirchevska2022reinforcement}, authors presented a novel Deep Q-Network (DQN) algorithm that considers the intention of other drivers. A method for learning intersection crossing policies is introduced in \cite{shalev2016safe}. The authors propose a hierarchical control architecture, which utilizes a high-level policy to choose the lane to use and a low-level policy to execute lane changes. Despite their great power in capturing complex interactions, pure RL-based approaches often struggle with safety assurance \cite{wang2020safe} due to the black box nature of these methods and the complexity of reward function design. Moreover, they can require vast amounts of training data, and their performance can be sensitive to changes in the environment or task \cite{liniger2019safe}.

\subsection{Hybrid Approaches}
Recognizing the complementary strengths and inherent limitations of MPC and RL, recent research efforts have begun to explore hybrid approaches that integrate these two methodologies. Some methods use a switching or supervisory mechanism between MPC and RL. For example, \cite{wang2020safe} and \cite{li2020deep} switch between MPC and RL controllers based on safety criteria, while \cite{bautista2022autonomous} run RL and MPC planners in parallel and select between them, and \cite{dang2023event} use RL to learn when MPC should recompute. Other studies adopt a more integrated architecture. \cite{liniger2019safe} introduced a hierarchical framework where RL is responsible for high-level decision-making while MPC performs trajectory optimization, and \cite{zhang2023learning} propose a framework in which RL tunes MPC parameters online, but the MPC handles the control mechanism solely. Brito et al. \cite{brito2022learning} proposed an Interactive Model Predictive Controller (IntMPC) in which a deep RL agent learns an interaction-aware velocity reference that is fed into the cost function of a Model Predictive Contour Controller (MPCC). They show superior performance of their hybrid framework over pure MPC.

Despite their valuable contributions, existing approaches have several limitations that our work addresses. First, their RL agent observes only two vehicles (the immediate leader and follower), which limits the interaction model to pairwise reasoning, whereas our framework handles simultaneous interactions with multiple surrounding vehicles from different directions and in different traffic densities. Second, they train and evaluate separate policies for each scenario and do not investigate the generalizability of their model, while we demonstrate zero-shot transfer of a learned policy to a structurally different environment. Third, their training procedure disables collision avoidance constraints in the optimization-based planner, then re-enables them at test time. This introduces a mismatch between training and deployment dynamics. In contrast, our framework maintains the MPC's collision avoidance mechanism throughout both training and evaluation, which ensures that the RL agent learns to operate in concert with the safety layer it will encounter during deployment.

\section{Methodology}
\subsection{Problem Formulation}
This section presents the mathematical formulation of the proposed framework. Consider an automated vehicle navigating through an unsignalized intersection characterized by a dynamical system with state space $\mathcal{X} \subseteq \mathbb{R}^n$ and control space $\mathcal{U} \subseteq \mathbb{R}^m$. The objective is to synthesize a control policy that ensures safe and efficient trajectory tracking while adhering to a predefined reference trajectory $\mathcal{T}_{\text{ref}}$ in the presence of stochastic agent behaviors. Let the reference trajectory be defined on a finite time horizon $[0,T]$ as:
\begin{equation}
    \mathcal{T}_{\text{ref}} = \{(x_{\text{ref},k}, y_{\text{ref},k}, v_{\text{ref},k}, \theta_{\text{ref},k})\}_{k=0}^N \in \mathcal{X}^N
\end{equation}

The control synthesis problem can be formulated as a constrained optimization problem subject to the following requirements: (1) trajectory tracking to minimize the tracking error, (2) safety assurance to maintain minimum safety distance from the set of adjacent vehicles, (3) input constraints to generate control inputs that satisfy actuator limitations and ensure smooth operation, and (4) adaptive behavior to modulate reference velocity profile based on environmental state and interaction patterns between vehicles. 

\begin{figure*}[t]
    \centering
    \includegraphics[width=0.75\linewidth]{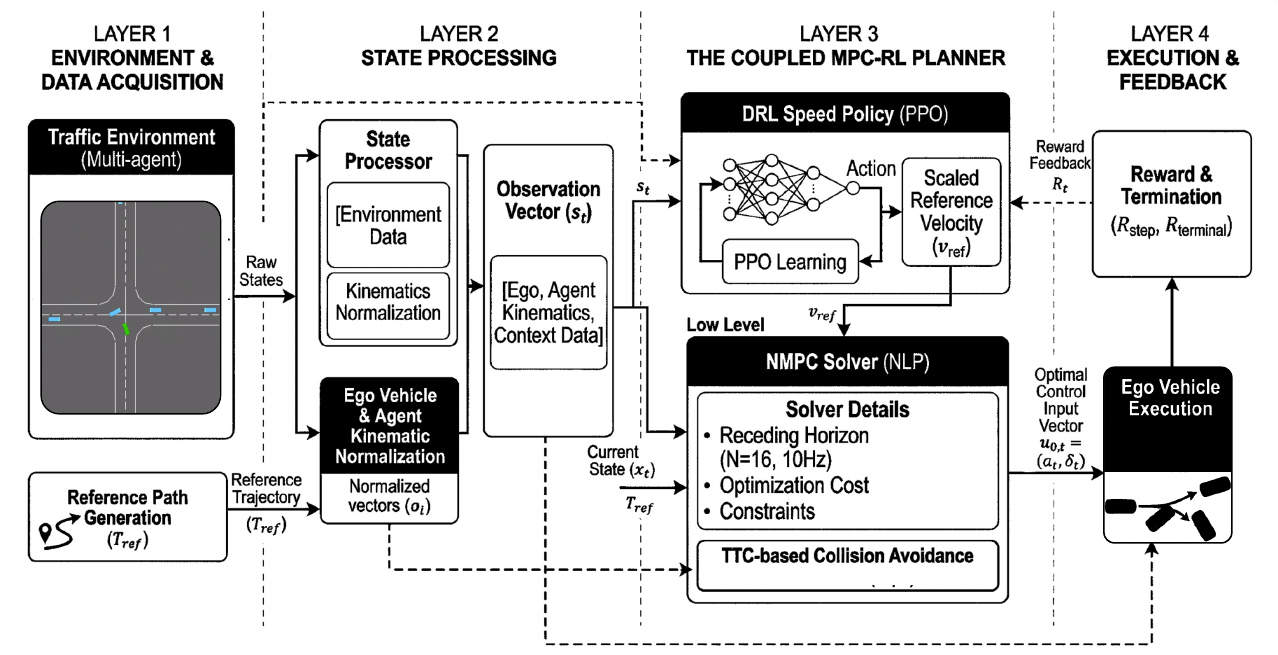}
    \caption{Proposed MPC-RL framework}
    \label{fig:framework}
\end{figure*}

To address this multi-objective control problem, we adopt an integrated MPC-RL framework that comprises an RL policy $\pi_\psi: \mathcal{S} \rightarrow \mathcal{V}$ that maps the environmental state space $\mathcal{S}$ to a feasible reference velocity space $\mathcal{V}$, parameterized by $\psi$, and an MPC controller that solves a receding horizon optimal control problem to generate control inputs $(a_k, \delta_k) \in \mathcal{U}$ while enforcing state and input constraints and tracking the RL-recommended reference speed. A schematic illustration of the proposed framework is presented in Fig.~\ref{fig:framework}.

\subsubsection{Vehicle Dynamics Model}
The vehicle's motion is represented using a kinematic bicycle model. The state vector at time $k$ is defined as $\mathbf{x}(k) = \begin{bmatrix} x_k,  y_k,  \theta_k, v_k \end{bmatrix}$, where $(x_k, y_k)$ represent the global coordinates, $\theta_k$ is the heading angle and $v_k$ is the velocity. The control input vector is $\mathbf{u}(k) = \begin{bmatrix} a_k, \delta_k \end{bmatrix}$, where $a_k$ is the longitudinal acceleration, $\delta_k$ is the steering angle.

The vehicle motion follows the continuous-time bicycle model dynamics:
\begin{equation}
    \begin{aligned}
        \dot{x} &= v \cos(\theta + \beta), \\
        \dot{y} &= v \sin(\theta + \beta), \\
        \dot{\theta} &= \frac{v}{L} \sin(\beta), \\
        \dot{v} &= a
    \end{aligned}
\end{equation}
where $\beta = \arctan\!\left(\frac{l_r}{L} \tan(\delta)\right)$ is the slip angle, $L$ is the wheelbase length, and $l_r$ is the distance from the rear axle to the center of mass.

\paragraph{Discretized Dynamics}

To facilitate real-time optimization, the continuous-time vehicle model is discretized using forward Euler integration with a sampling time \( T_s \). The full nonlinear dynamics, including the slip angle \( \beta \), are retained in the discretized formulation:
\begin{equation}
    \mathbf{x}_{k+1} = \mathbf{x}_k + f_c(\mathbf{x}_k, \mathbf{u}_k)\, T_s
\end{equation}
where \( f_c(\cdot, \cdot) \) denotes the continuous-time dynamics in Eq.~(2). The resulting discrete-time update equations are:
\begin{equation}
    \begin{aligned}
        x_{k+1} &= x_k + v_k \cos(\theta_k + \beta_k)\, T_s \\
        y_{k+1} &= y_k + v_k \sin(\theta_k + \beta_k)\, T_s \\
        \theta_{k+1} &= \theta_k + \frac{v_k}{L} \sin(\beta_k)\, T_s \\
        v_{k+1} &= v_k + a_k\, T_s
    \end{aligned}
\end{equation}
where \( \beta_k = \arctan\!\left(\frac{l_r}{L} \tan(\delta_k)\right) \). For compactness, we denote the corresponding discrete-time transition map by
\begin{equation}
    \mathbf{x}_{k+1} = f_d(\mathbf{x}_k, \mathbf{u}_k).
\end{equation}
Because the dynamics are nonlinear, the resulting MPC optimization is formulated as a nonlinear program (NLP) and solved at each control step using the interior-point solver IPOPT \cite{wachter2006implementation} within the CasADi framework \cite{andersson2019casadi}.

\subsubsection{Model Predictive Control (MPC) Formulation}

The MPC controller solves a finite-horizon constrained optimization problem at each time step to ensure trajectory tracking, kinematic feasibility, and smooth motion.

\paragraph{Cost Function}

The cost function over a prediction horizon \( N \) penalizes state tracking error, control effort, input variation, and terminal deviation:
\begin{align}
    J = \sum_{k=0}^{N-1} & \left( \| e_{XY,k}^\perp \|^2_{Q_{XY,\perp}} + \| e_{XY,k}^\parallel \|^2_{Q_{XY,\parallel}} + \| e_{\theta v, k} \|^2_{Q_{\theta v}} \right) \notag \\
    & + \sum_{k=0}^{N-1} \left( \| u(k) \|^2_R + \| u(k+1) - u(k) \|^2_{R_d} \right) \notag \\
    & + \| x_N - x_{\text{ref}, N} \|^2_{Q_f}
\end{align}
Here, the position error $e_{XY,k} = x_{XY}(k) - x_{\text{ref},XY}(k)$ is decomposed into lateral and longitudinal components relative to the unit tangent vector $\hat{\tau}_k$ of the reference path: $e_{XY,k}^\perp$ captures perpendicular deviation and $e_{XY,k}^\parallel$ captures along-path deviation. The term $e_{\theta v,k}$ represents yaw and velocity deviation from the reference. The matrices $Q_{XY,\perp}$, $Q_{XY,\parallel}$, $Q_{\theta v}$, $R$, $R_d$, and $Q_f$ are weighting matrices that balance the relative importance of each cost component.

\paragraph{Constraints}

The optimization is subject to the discretized nonlinear dynamics constraint $\mathbf{x}_{k+1} = f_d(\mathbf{x}_k, \mathbf{u}_k)$ for all $k \in [0, N{-}1]$, and bounded control inputs $a_{\min} \leq a_k \leq a_{\max}$, $\delta_{\min} \leq \delta_k \leq \delta_{\max}$.

\subsubsection{Collision Avoidance}
In our implementation, collision avoidance is handled through a TTC-triggered modification of the reference velocity rather than through an explicit hard collision-avoidance constraint in the MPC optimization problem. For each surrounding vehicle $i$, future positions are predicted using a constant-velocity model: $x_{i,k+1} = x_{i,k} + v_{i,k}\cos(\theta_{i,k})\Delta t$, $y_{i,k+1} = y_{i,k} + v_{i,k}\sin(\theta_{i,k})\Delta t$. Potential collisions are detected by finding intersection points between the ego vehicle's planned trajectory and these predicted trajectories, and the time-to-collision is computed as:
\begin{equation}
    \text{TTC}_{i} = \min_{k} \{\Delta t \cdot k \mid d((x_k, y_k), (x_{\text{int}}, y_{\text{int}})) < d_{\min}\}
\end{equation}
where $d(\cdot,\cdot)$ is the Euclidean distance and $d_{\min}$ is a proximity threshold. When $\text{TTC} < \text{TTC}_{\text{threshold}}$, the reference velocity is replaced with a linear deceleration profile $v_{\text{ref},k} = v_{\text{current}} \cdot (1 - k/k_{\text{stop}})$ for $k < k_{\text{stop}}$ and zero thereafter, where $k_{\text{stop}}$ is the number of steps to reach zero velocity.

\subsection{Reinforcement Learning Component}

The RL component operates at a higher level in the control hierarchy, determining the reference speed that the MPC controller should track.

\subsubsection{RL Problem Formulation}

The speed adaptation problem is formulated as a Markov Decision Process (MDP) where the RL agent learns to generate appropriate reference speeds based on the traffic situation.

\paragraph{State Space}
The state observation provided to the RL agent at each timestep consists of normalized kinematic information for both the ego vehicle and surrounding vehicles, augmented with MPC-context features. The raw kinematic feature vector for each vehicle $i$ is:
\begin{equation}
    \mathbf{o}_i = [p_i, x_i, y_i, v_{x,i}, v_{y,i}, \theta_i, \sin(\theta_i), \cos(\theta_i)]
\end{equation}
where $p_i$ is a presence indicator, $(x_i, y_i)$ are global coordinates, $(v_{x,i}, v_{y,i})$ are velocity components, and $\theta_i$ is the heading angle, with $i=0$ denoting the ego vehicle and $N_v$ the maximum number of observable surrounding vehicles. Each feature is normalized to approximately $[-1, 1]$ using predefined ranges. The normalized and flattened kinematic vectors are concatenated with six MPC-context features to form the complete state vector:
\begin{equation}
    s_t = [\text{flatten}(\{\bar{\mathbf{o}}_i\}_{i=0}^{N_v}),\; \tilde{v}_{\text{ego}},\; \rho,\; d_{\min},\; d_{\text{conflict}},\; n_{\text{vis}},\; t_{\text{rem}}]
\end{equation}
where $\bar{\mathbf{o}}_i$ denotes the normalized kinematic vector, $\tilde{v}_{\text{ego}} = v_{\text{ego}} / v_{\text{max}}$ is the normalized ego speed, $\rho \in [0,1]$ is the progress along the reference trajectory, $d_{\min} \in [0,1]$ is the normalized minimum distance to the nearest vehicle, $d_{\text{conflict}} \in [0,1]$ is the normalized distance to the nearest predicted conflict point, $n_{\text{vis}}$ is the normalized count of visible vehicles, and $t_{\text{rem}} \in [0,1]$ is the fraction of episode time remaining. With $N_v = 9$ and 8 features per vehicle (including the ego), this yields an 86-dimensional observation vector (80 kinematic + 6 context features).

\paragraph{Action Space}
The RL agent outputs a normalized reference speed multiplier:
\begin{equation}
    \alpha_t \in [-1, 1]
\end{equation}
which is then scaled to an actual reference speed $v_{\text{ref}} = v_{\text{max}} \cdot \frac{(\alpha_t + 1)}{2}$, where $v_{\text{max}}$ is the maximum allowable speed.

\paragraph{Reward Function}
The reward function combines terminal rewards for episode-ending events with per-step incentives:
\begin{equation}
    R(s_t, \alpha_t) = \begin{cases}
        R_{\text{crash}} = -50 & \text{if collision} \\
        R_{\text{off-road}} = -10 & \text{if off-road departure} \\
        R_{\text{arrival}} = +20 & \text{if reached destination} \\
        R_{\text{step}} & \text{otherwise}
    \end{cases}
\end{equation}
The per-step reward $R_{\text{step}} = R_{\text{alive}} + R_{\text{speed}} + R_{\text{proximity}} + R_{\text{smooth}} + R_{\text{center}}$ is composed of five terms detailed in Table~\ref{tab:reward_components}.

\begin{table}[h]
    \centering
    \caption{Per-step reward components.}
    \label{tab:reward_components}
    \begin{tabular}{lll}
        \toprule
        \textbf{Term} & \textbf{Formula} & \textbf{Role} \\
        \midrule
        $R_{\text{alive}}$     & $0.2$                                          & Survival bonus \\
        $R_{\text{speed}}$     & $1.5 \cdot \min(v_k / v_{\max},\, 1.5)$       & Progress \\
        $R_{\text{proximity}}$ & $-2.0(1 - 2d_{\min})^2$ if $d_{\min} < 0.5$   & Safety \\
        $R_{\text{smooth}}$    & $-0.3 \cdot |\Delta a_k| / a_{\max}$          & Comfort \\
        $R_{\text{center}}$    & $0.3\left(1 - \frac{|l_k|}{w_{\text{lane}}/2}\right)$ & Lane keeping \\
        \bottomrule
    \end{tabular}
\end{table}

\subsubsection{Policy Learning}
The RL policy is trained using Proximal Policy Optimization (PPO) with the following objective:
\begin{equation}
    L^{\text{PPO}}(\psi) = \mathbb{E}_t[\min(\rho_t(\psi)\hat{A}_t, \text{clip}(\rho_t(\psi), 1-\epsilon_{\text{clip}}, 1+\epsilon_{\text{clip}})\hat{A}_t)]
\end{equation}
, where $\rho_t(\psi)$ is the probability ratio between new and old policies, $\hat{A}_t$ is the estimated advantage function, and $\epsilon_{\text{clip}}$ is the clipping parameter.

\subsection{MPC-RL Integration}

The control loop operates as follows: At each timestep $t$, the RL policy observes the current state $s_t$ and outputs a reference speed $v_{\text{ref}}$, and the MPC controller solves the optimization problem:
    \begin{equation}
        \min_{\mathbf{u}_{0:N-1}} J_{\text{total}} \quad \text{subject to:}
    \end{equation}
    \begin{align}
        \mathbf{x}_{k+1} &= f_d(\mathbf{x}_k, \mathbf{u}_k) \\
        \mathbf{x}_k &\in \mathcal{X}, \mathbf{u}_k \in \mathcal{U} \\
        v_k &\approx v_{\text{ref}} \text{ (soft constraint)}
    \end{align}
    
The first control input $\mathbf{u}_0$ is applied to the vehicle. During the training phase, the environment returns a reward value to the RL agent based on the action taken by the vehicle (output of the MPC controller). The process repeats with the updated state.

\section{Experiments and Results}

To evaluate the effectiveness of the proposed MPC-RL framework, we conduct
a series of experiments as explained in the following sections.

\subsection{Experimental Setup} Our experimental evaluation was conducted in the Highway-Env simulation environment~\cite{highway-env}. The environment features dynamic traffic conditions with multiple interacting vehicles exhibiting stochastic behaviors, which makes it suitable for evaluating decision-making and control strategies. To systematically assess how each method scales with increasing traffic complexity, we evaluate across three difficulty levels, with 1,000 randomly initiated episodes per method per difficulty level, which yields 3,000 episodes per method in total. Difficulty is controlled by two parameters: \textit{initial\_vehicle\_count}, the number of vehicles present when the episode begins, and \textit{spawn\_probability}, the probability that a new vehicle spawns at each timestep, so that higher values produce continuously arriving traffic. The three levels are Easy (2~vehicles, spawn probability~0.1), Moderate (5~vehicles, spawn probability~0.3), and Hard (10~vehicles, spawn probability~0.6). Importantly, at all difficulty levels, the surrounding vehicles approach the intersection from multiple directions with stochastic behaviors, which creates scenarios where conflicts are frequent, and some collisions are difficult to avoid regardless of the ego vehicle's control strategy. This is intentional to provide a highly adversarial test bed for our models. It is also worth noting that MPC-RL is only trained in hard traffic conditions, and evaluation in other traffic conditions is without retraining the model. Additionally, we evaluate the transferability of the trained MPC-RL policy to a highway merging scenario without any retraining to assess its zero-shot generalization capability. The merging evaluation mirrors the intersection protocol with 1,000 episodes per method per difficulty level across three traffic densities: Easy (1~vehicle), Moderate (2~vehicles), and Hard (4~vehicles).

We implemented and compared three approaches to evaluate the performance of the proposed framework:
\begin{itemize}
\item \textbf{Pure MPC}: A baseline MPC implementation with manual collision avoidance described in Section~III.
\item \textbf{Pure PPO}: A standalone PPO implementation without MPC integration, trained end-to-end for both decision-making and control.
\item \textbf{MPC-RL}: Our proposed framework that combines PPO-based decision-making with MPC-based trajectory optimization and control.
\end{itemize}


The evaluation focuses on three key metrics: success rate (arriving at the destination safely), collision rate, and other outcomes (off-road departures and timeouts combined). Episodes that do not end in success, collision, or off-road departure are classified as timeouts, which occur when the vehicle fails to complete the maneuver within the allotted episode length (130 steps, corresponding to 13 seconds at 10 Hz control frequency). Key training and MPC hyperparameters are summarized in Table~\ref{tab:hyperparams}.

\begin{table}[h]
    \centering
    \caption{Training and MPC hyperparameters}
    \label{tab:hyperparams}
    \begin{tabular}{ll}
        \hline
        \textbf{Parameter} & \textbf{Value} \\
        \hline
        Training budget & 4,000,000 steps \\
        Batch size / Minibatch size & 4,096 / 64 \\
        PPO epochs per update & 10 \\
        Learning rate / Discount factor & $10^{-4}$ / 0.99 \\
        Policy network & FC [512, 256] \\
        MPC prediction horizon & 16 steps \\
        Control frequency & 10 Hz \\
        Nominal reference speed & 10 m/s \\
        \hline
    \end{tabular}
\end{table}

\subsection{Intersection Performance Analysis}
\begin{figure}
    \centering
    \includegraphics[width=0.95\linewidth]{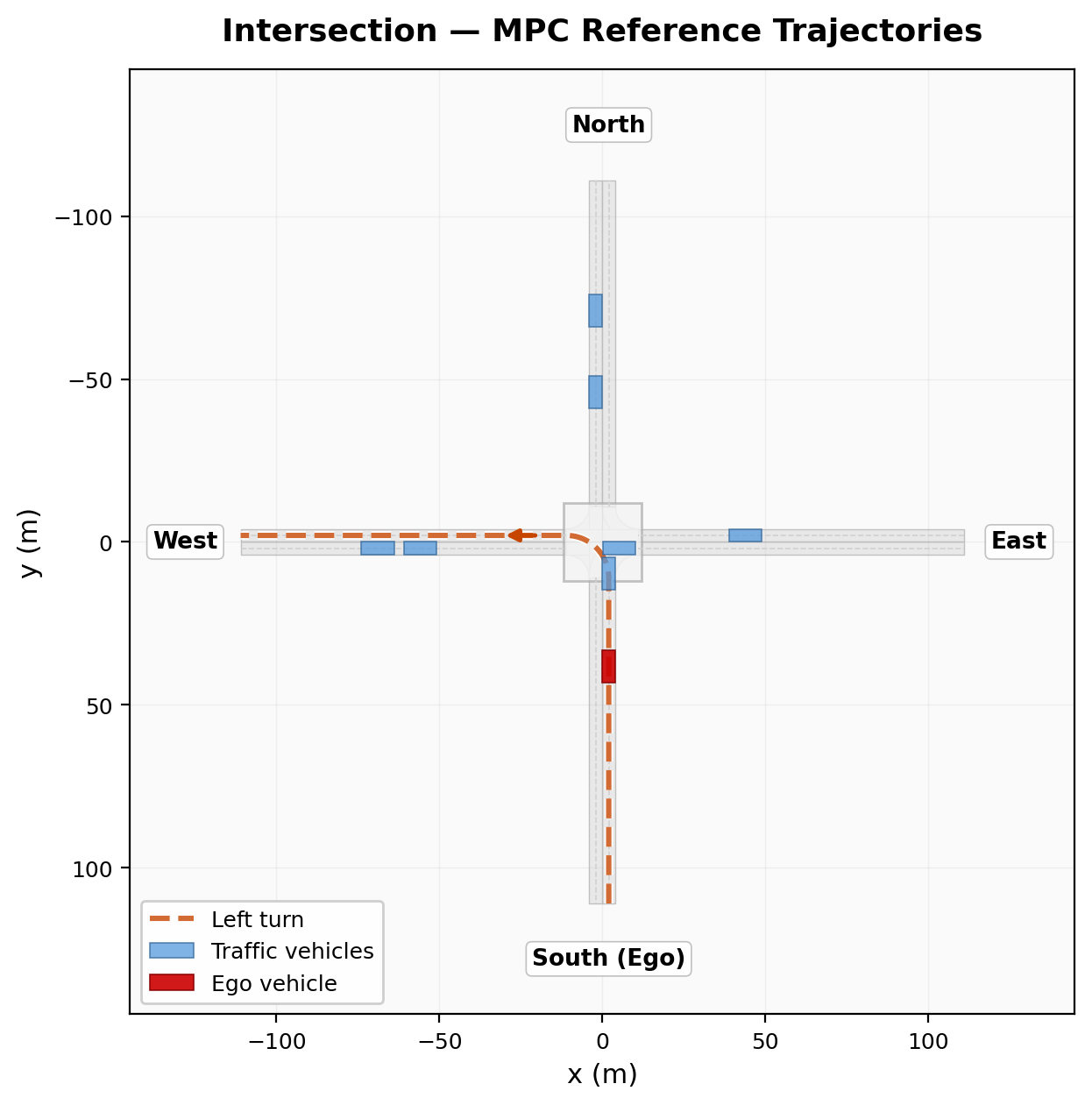}
    \caption{Intersection scenario with ego vehicle and MPC reference trajectory.}
    \label{fig:intersection}
\end{figure}

Fig.~\ref{fig:intersection} illustrates a snapshot of the intersection scenario that shows the ego vehicle (red) approaching from the south with its planned left-turn reference trajectory alongside surrounding traffic vehicles arriving from multiple directions. 

Table~\ref{tab:intersection_results} presents the performance of each method across the three difficulty levels along with pooled results. MPC-RL achieves the highest success rate and lowest collision rate at every difficulty level. The advantage is statistically significant for both metrics in the Easy ($p < 0.001$) and Moderate ($p < 0.01$) conditions; in the Hard condition, the success rate difference remains significant ($p = 0.039$) while the collision rate difference approaches but does not reach significance ($p = 0.053$). This suggests that the benefit of RL-guided speed adaptation attenuates in extremely conflict-dense traffic. As the difficulty increases, all methods exhibit degraded performance, as expected given the denser traffic and more frequent conflicts. However, MPC-RL consistently maintains an advantage over both baselines. 

\begin{table}[t]
    \centering
    \caption{Intersection performance by difficulty level (1,000 episodes per method per level) and pooled across all levels (3,000 episodes per method), with pairwise statistical comparisons on the pooled data.}
    \label{tab:intersection_results}
    \begin{tabular}{llccc}
        \toprule
        \textbf{Difficulty} & \textbf{Method} & \begin{tabular}[c]{@{}c@{}}\textbf{Success}\\\textbf{(\%)}\end{tabular} & \begin{tabular}[c]{@{}c@{}}\textbf{Collision}\\\textbf{(\%)}\end{tabular} & \begin{tabular}[c]{@{}c@{}}\textbf{Other}\\\textbf{(\%)}\end{tabular} \\
        \midrule
        \multirow{3}{*}{Easy}
            & \textbf{MPC-RL} & \textbf{94.8} & \textbf{5.2} & 0.0 \\
            & Pure MPC        & 89.4          & 10.5         & 0.1 \\
            & Pure PPO        & 83.4          & 5.9          & 10.7 \\
        \midrule
        \multirow{3}{*}{Moderate}
            & \textbf{MPC-RL} & \textbf{82.6} & \textbf{17.2} & 0.2 \\
            & Pure MPC        & 77.6          & 22.3          & 0.1 \\
            & Pure PPO        & 73.8          & 26.1          & 0.1 \\
        \midrule
        \multirow{3}{*}{Hard}
            & \textbf{MPC-RL} & \textbf{67.3} & \textbf{32.2} & 0.5 \\
            & Pure MPC        & 62.8          & 36.4          & 0.8 \\
            & Pure PPO        & 53.3          & 42.1          & 4.6 \\
        \midrule
        \multirow{3}{*}{Pooled}
            & \textbf{MPC-RL} & \textbf{81.6} & \textbf{18.2} & 0.2 \\
            & Pure MPC        & 76.6          & 23.1          & 0.3 \\
            & Pure PPO        & 70.2          & 24.7          & 5.1 \\
        \midrule
        \multicolumn{5}{l}{\textit{Pairwise comparisons on pooled data (Fisher's exact test):}} \\
        \midrule
        \multicolumn{2}{l}{\textbf{Comparison}} & $\Delta$ Success & $\Delta$ Collision & \\
        \midrule
        \multicolumn{2}{l}{MPC-RL vs.\ Pure MPC} & +6.5\%***  & $-21.2\%$*** & \\
        \multicolumn{2}{l}{MPC-RL vs.\ Pure PPO} & +16.2\%*** & $-26.3\%$*** & \\
        \multicolumn{2}{l}{Pure MPC vs.\ Pure PPO} & +9.1\%*** & $-6.5\%$\textsuperscript{ns} & \\
        \bottomrule
        \multicolumn{5}{l}{\footnotesize Other = off-road departures + timeouts. $\Delta$ = relative change.} \\
        \multicolumn{5}{l}{\footnotesize *** $p<0.001$; \textsuperscript{ns} not significant ($p=0.146$).}
    \end{tabular}
\end{table}

The pooled results confirm that MPC-RL significantly outperforms Pure MPC in success rate by $+6.5\%$ and collision rate by $-21.2\%$. MPC-RL also significantly outperforms Pure PPO on both metrics ($+16.2\%$ success; $-26.3\%$ collision). Interestingly, while Pure MPC significantly outperforms Pure PPO in success rate ($+9.1\%$, $p < 0.001$), the difference in collision rate between Pure MPC and Pure PPO is not statistically significant ($-6.5\%$, $p = 0.146$). 

Overall, the pure MPC controller, relying solely on its hand-crafted TTC-based collision avoidance strategy, tends to either react too conservatively (leading to timeouts) or misjudge the timing of other vehicles' trajectories, which results in a significantly higher collision rate than MPC-RL. In contrast, the MPC-RL framework leverages the RL agent's learned understanding of multi-agent interaction dynamics to provide more informed speed references, which enables the MPC controller to execute trajectories that better balance assertiveness with caution. The simultaneous improvement in both success rate and collision rate across all difficulty levels indicates that the RL component learned to generate speed references that help the vehicle navigate the intersection more decisively by reducing the number of episodes that end in either collision or timeout.

The pure PPO approach demonstrated the poorest overall performance with the lowest success rate across all difficulty levels and the only substantial other-outcome rates among all tested methods. Also, it suffers from a relatively high rate of other outcomes (primarily off-road departures), which reflects its inability to maintain feasible trajectories without explicit kinematic constraints. 
This highlights the fundamental challenge of learning both high-level decision-making and low-level control simultaneously. We note that more extensive reward engineering or a substantially larger training budget could potentially improve pure PPO's performance; however, our comparison deliberately uses the same reward structure and a common training budget across both learning-based methods to isolate the architectural advantages or disadvantages of different designs. 

It is important to note that the high collision rates observed, particularly in the Hard condition, are a consequence of the intentionally adversarial nature of the evaluation scenarios, where other vehicles do not avoid collisions and their spawning and initialization are deliberately configured to create challenging conflict situations, including cases in which some collisions are difficult or impossible to avoid.

\subsection{Zero-Shot Transfer to Highway Merging}

\begin{figure}[b]
    \centering
    \includegraphics[width=0.95\linewidth]{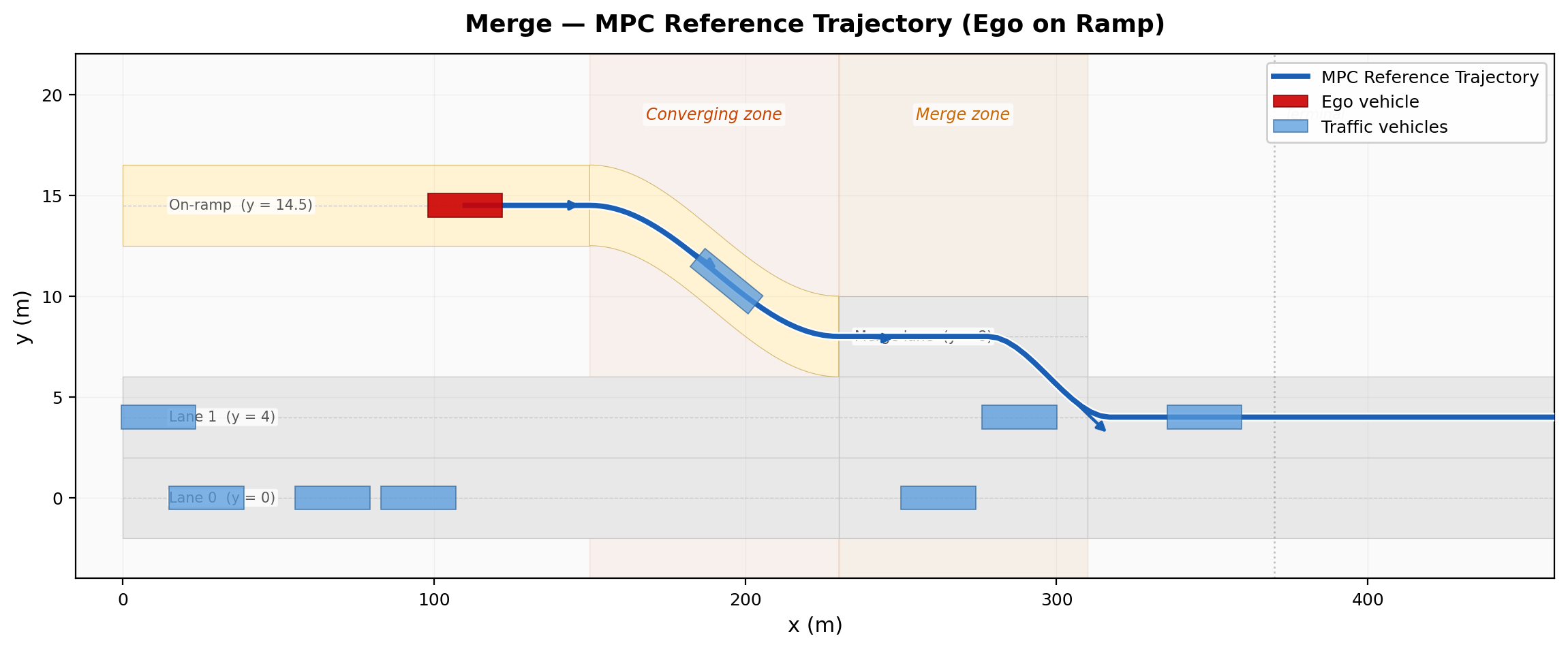}
    \caption{Illustration of merging experiment with ego vehicle merging into the highway.}
    \label{fig:merge}
\end{figure}

To assess the generalization capability of the proposed framework, we evaluate all three methods on a highway merging scenario without any retraining of the RL policy. This experiment tests whether the RL agent's learned understanding of multi-agent interactions (acquired exclusively in the intersection environment) can transfer to a structurally different driving scenario. In the merging scenario, the ego vehicle must join a stream of highway traffic from an on-ramp, which requires assessing gaps, adjusting speed, and negotiating with vehicles on the main road (Fig.~\ref{fig:merge}). While the geometric layout and traffic flow patterns differ substantially from the intersection setting, the underlying decision-making challenge shares similarities: the ego vehicle must determine when to proceed assertively and when to yield, based on the relative positions and velocities of surrounding vehicles. As with the intersection evaluation, we test across three difficulty levels with 1,000 episodes per method per level.

The results in Table~\ref{tab:merge_comparison} reveal several important findings. First, pure PPO fails completely across all difficulty levels, with 100\% off-road departures within the first few simulation steps. This demonstrates that an intersection-trained end-to-end RL policy shows zero transferability to the merge geometry without the MPC backbone, as the learned low-level control commands are entirely unsuitable for the new road layout.

\begin{table}[t]
    \centering
    \caption{Highway merging performance by difficulty level (1,000 episodes per method per level; zero-shot transfer for MPC-RL and Pure PPO).}
    \label{tab:merge_comparison}
    \begin{tabular}{llcccc}
        \toprule
        \textbf{Difficulty} & \textbf{Method} & \begin{tabular}[c]{@{}c@{}}\textbf{Success}\\\textbf{(\%)}\end{tabular} & \begin{tabular}[c]{@{}c@{}}\textbf{Collision}\\\textbf{(\%)}\end{tabular} & \begin{tabular}[c]{@{}c@{}}\textbf{Off-road}\\\textbf{(\%)}\end{tabular} & \textbf{Sig.} \\
        \midrule
        \multirow{3}{*}{Easy}
            & MPC-RL   & \textbf{100.0} & \textbf{0.0}  & 0.0   & \textsuperscript{ns} \\
            & Pure MPC & \textbf{100.0} & \textbf{0.0}  & 0.0   & \\
            & Pure PPO & 0.0            & 0.0           & 100.0 & \\
        \midrule
        \multirow{3}{*}{Moderate}
            & MPC-RL   & \textbf{100.0} & \textbf{0.0}  & 0.0   & \textsuperscript{ns} \\
            & Pure MPC & \textbf{100.0} & \textbf{0.0}  & 0.0   & \\
            & Pure PPO & 0.0            & 0.0           & 100.0 & \\
        \midrule
        \multirow{3}{*}{Hard}
            & MPC-RL   & 83.3           & 16.7          & 0.0   & ** \\
            & Pure MPC & \textbf{87.8}  & \textbf{12.2} & 0.0   & \\
            & Pure PPO & 0.0            & 0.0           & 100.0 & \\
        \bottomrule
        \multicolumn{6}{l}{\footnotesize Sig.\ = Fisher's exact test for MPC-RL vs.\ Pure MPC.} \\
        \multicolumn{6}{l}{\footnotesize ** $p<0.01$; \textsuperscript{ns} not significant (identical outcomes).}
    \end{tabular}
\end{table}
Second, in the Easy and Moderate conditions, both MPC-RL and pure MPC achieve a perfect 100\% success rate with zero collisions or off-road departures, confirming that both MPC-based methods handle the merge maneuver flawlessly under low to moderate traffic density. The two approaches are statistically indistinguishable at these levels.

Third, in the Hard condition (4~vehicles), the pattern reverses relative to the intersection results. Pure MPC outperforms MPC-RL by achieving 87.8\% success and 12.2\% collision versus 83.3\% success and 16.7\% collision for MPC-RL. This reversal is informative. The RL policy, trained exclusively on intersection dynamics, has learned speed modulation strategies tailored to multi-directional conflict resolution. When applied to the structurally different merging geometry, these intersection-specific speed adjustments introduce perturbations that can interfere with the MPC solver's ability to identify optimal merge windows. 

Despite the reversal at the Hard level, these results yield two encouraging findings for the framework's generalization potential. Both MPC-based methods achieve strong absolute performance even at the hardest merging condition (83.3--87.8\% success in a completely unseen scenario), demonstrating that the MPC backbone provides a robust foundation for cross-scenario transfer. Furthermore, the RL component does not catastrophically degrade the system's behavior, as MPC-RL maintains zero off-road departures and a reasonable collision rate, in contrast to the complete failure of pure PPO. 

\subsection{Training Efficiency}
\begin{figure}[h]
    \centering
    \includegraphics[width=0.9\linewidth]{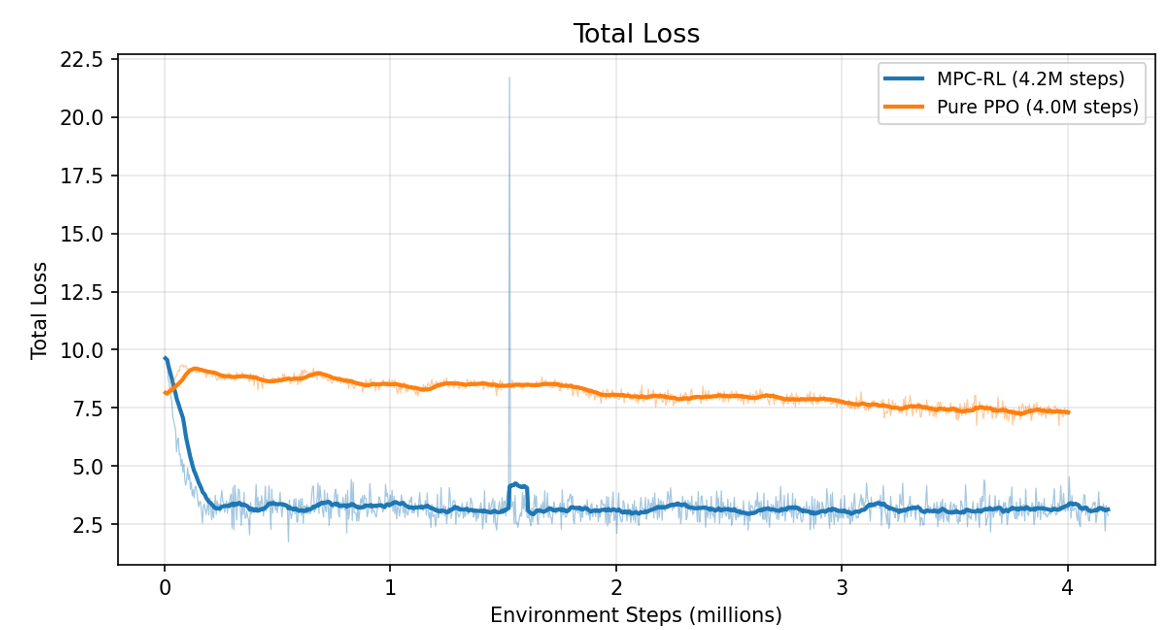}
    \caption{Training progress: Loss values for pure RL (PPO) and MPC-RL models.}
    \label{fig:training_convergence}
\end{figure}

Fig.~\ref{fig:training_convergence} compares the training loss evolution of MPC-RL and pure PPO. MPC-RL converges rapidly, with its total loss dropping from approximately 9.5 to around 3.0 within the first 200,000 steps and stabilizing thereafter. In contrast, pure PPO's loss remains above 7.0 for most of training and only reaches approximately 6.5 after 2.5 million steps, reflecting the difficulty of learning both decision-making and low-level control simultaneously. We note that the two agents operate on different action spaces, so their absolute loss values are not directly comparable; however, the relative comparison, together with the observation that MPC-RL achieves superior task performance (Table~\ref{tab:intersection_results}), suggests that restricting the RL agent to high-level speed adaptation reduces the optimization burden during training.

\subsection{Computational Cost}
To assess the real-time feasibility of the proposed framework, we measure per-step inference time for each method on the intersection scenario. Table~\ref{tab:runtime} reports the mean, standard deviation, and median wall-clock time per control step. Pure PPO is the fastest method (mean 0.71\,ms), as it requires only a single forward pass through the policy network. Pure MPC requires 38.04\,ms on average, dominated by the NLP solve via IPOPT. The MPC-RL framework has a total mean inference time of 38.36\,ms, which represents less than 1\% overhead relative to pure MPC, as the RL forward pass is negligible compared to the MPC solve. All methods comfortably satisfy the 100\,ms budget imposed by the 10\,Hz control frequency.

\begin{table}[h]
    \centering
    \caption{Per-step inference time on the intersection scenario.}
    \label{tab:runtime}
    \begin{tabular}{lccc}
        \toprule
        \textbf{Method} & \begin{tabular}[c]{@{}c@{}}\textbf{Mean}\\\textbf{(ms)}\end{tabular} & \begin{tabular}[c]{@{}c@{}}\textbf{Std}\\\textbf{(ms)}\end{tabular} & \begin{tabular}[c]{@{}c@{}}\textbf{Median}\\\textbf{(ms)}\end{tabular} \\
        \midrule
        Pure PPO & 0.71  & 0.09  & 0.70  \\
        Pure MPC & 38.04 & 13.72 & 35.95 \\
        MPC-RL   & 38.36 & 10.06 & 36.98 \\
        \bottomrule
    \end{tabular}
\end{table}

\section{Discussion and Limitations}
The experimental results suggest that the MPC-RL framework offers a promising direction for balancing safety and efficiency in automated intersection navigation. Across all difficulty levels, MPC-RL consistently achieves the highest success rate and lowest collision rate, indicating that RL-guided speed adaptation provides clear benefits beyond hand-crafted TTC-based collision avoidance alone. The advantage is most pronounced in lower-density traffic.
The zero-shot transfer results present a nuanced picture. At lower traffic densities, both MPC-based methods achieve perfect merging performance, while at the highest density, pure MPC outperforms MPC-RL. However, end-to-end PPO fails entirely in transfer, highlighting that the MPC backbone is the primary enabler of cross-scenario robustness. The hierarchical architecture appears to mitigate catastrophic transfer failure: even when the RL component provides suboptimal speed references in a new domain, the MPC layer preserves substantially better behavior than end-to-end PPO. Moreover, learning adaptive speed references from multi-vehicle interactions proves far more transferable than learning raw control commands in an end-to-end fashion.
The balanced behavior demonstrated by our framework also has potential implications for social acceptance of automated vehicles, as previous research has shown that overly conservative driving styles can lead to unsafe or inefficient situations \cite{rahmani2025automated, lee2021assessing}. Despite these findings, extensive real-world validation would be needed before making stronger claims about practical deployment.
Several limitations should be acknowledged. First, our baselines are limited to standalone MPC and end-to-end PPO; comparisons against other hybrid approaches or safety-constrained RL methods would provide a more complete picture. Second, Highway-Env uses simplified vehicle dynamics and idealized road geometries, so transferability to real-world conditions remains open. Third, ablation studies are needed to isolate the RL component's contribution relative to the TTC-based collision avoidance. Finally, the performance reversal in hard merging suggests that domain-specific adaptation strategies merit further investigation.

\section{Conclusion}
This study presents a coupled MPC-RL framework for automated navigation at unsignalized intersections, in which the RL agent learns adaptive speed references from multi-vehicle interactions while MPC handles trajectory optimization and constraint enforcement. Experimental results across three traffic-density levels show that MPC-RL consistently outperforms both standalone MPC and end-to-end PPO by achieving statistically significant improvements in both success rate and collision rate. The advantage is consistent across difficulty levels, though it becomes smaller in the most conflict-dense scenarios. Zero-shot transfer experiments to highway merging reveal that both MPC-based methods remain effective in the unseen scenario, whereas end-to-end PPO fails. This suggests that the MPC backbone can facilitate cross-domain generalization. Importantly, the hybrid architecture helps mitigate catastrophic failure even when the RL component operates out of distribution. The framework also shows faster loss stabilization during training than end-to-end RL, which is consistent with the idea that restricting the RL agent’s role reduces the learning burden. Future work should focus on broader baseline comparisons, including safety-constrained RL methods, domain adaptation strategies for improved transfer, ablation studies to isolate individual component contributions, and evaluation in higher-fidelity simulators.

	
	\bibliographystyle{IEEEtran}
	\bibliography{root} 
	
\end{document}